%% file: sample-sigconf.tex
\documentclass[sigconf]{acmart}
\AtBeginDocument{%
  }

\setcopyright{acmlicensed}
\copyrightyear{2026}
\acmYear{2026}
\acmDOI{XXXXXXX.XXXXXXX}
\acmConference[Conference acronym 'XX]{Make sure to enter the correct
  conference title from your rights confirmation email}{June 03--05,
  2018}{Woodstock, NY}
\acmISBN{978-1-4503-XXXX-X/2018/06}

\usepackage{enumitem}

\begin{document}
\title{LOLGORITHM: Funny Comment Generation Agent For Short Videos}

\author{Xuan Ouyang}
\affiliation{%
  \institution{University of New South Wales}
  \city{Sydney}
  \country{Australia}
}
\email{xuan.ouyang@unsw.edu.au}

\author{Bozhou Wang}
\affiliation{%
  \institution{University of Sydney}
  \city{Sydney}
  \country{Australia}
}
\email{bwan0613@uni.sydney.edu.au}

\author{Senan Wang}
\affiliation{%
  \institution{University of Sydney}
  \city{Sydney}
  \country{Australia}
}
\email{swan0472@uni.sydney.edu.au}

\author{Siyuan Xiahou}
\affiliation{%
  \institution{The University of Hong Kong}
  \city{Hong Kong}
  \country{China}
}
\email{u3634276@connect.hku.hk}

\author{Jinrong Zhou}
\affiliation{%
  \institution{University of Southern California}
  \city{Los Angeles}
  \country{USA}
}
\email{jinrongz@usc.edu}

\author{Yuekang Li}
\affiliation{%
  \institution{University of New South Wales}
  \city{Sydney}
  \country{Australia}
}
\email{yuekang.li@unsw.edu.au}










\begin{abstract}
Short-form video platforms have become central to multimedia information dissemination, where comments play a critical role in driving engagement, propagation, and algorithmic feedback. However, existing approaches — including video summarization and live-streaming danmaku generation — fail to produce authentic comments that conform to platform-specific cultural and linguistic norms. In this paper, we present LOLGORITHM, a novel modular multi-agent framework for stylized short-form video comment generation. LOLGORITHM supports six controllable comment styles and comprises three core modules: video content summarization, video classification, and comment generation with semantic retrieval and hot meme augmentation. We further construct a bilingual dataset of 3,267 videos and 16,335 comments spanning five high-engagement categories across YouTube and Douyin. Evaluation combining automatic scoring and large-scale human preference analysis demonstrates that LOLGORITHM consistently outperforms baseline methods, achieving human preference selection rates of 80.46\% on YouTube and 84.29\% on Douyin across 107 respondents. Ablation studies confirm that these gains are attributable to the framework architecture rather than the choice of backbone LLM, underscoring the robustness and generalizability of our approach.
\end{abstract}




\settopmatter{printfolios=true}
\maketitle

\input{input/introduction}
\input{input/related_work}
\input{input/methodology}
\input{input/evaluation}
\input{input/conclusion}
\bibliographystyle{ACM-Reference-Format}
\bibliography{references}










\end{document}

%% file: input/introduction.tex
\section{Introduction}
Short video platforms have become crucial channels for multimedia information dissemination, where comments play a vital role in content propagation, community interaction, and algorithmic feedback. With the development of generative AI, comments are transforming from auxiliary information to core driving forces in platform ecosystems.

Despite the remarkable capabilities of current Multimodal Large Language Models (MLLMs), and the fact that iterative model improvements have enabled them to generate video summaries that structurally resemble comments through prompt engineering alone, they still fall short of producing genuine comments. The generated text remains essentially a summary of the video content, failing to meet the fundamental requirements of authentic comments. Meanwhile, existing alternative approaches either focus on video summarization---which lacks stylistic expressiveness---or target live-streaming danmaku generation, which cannot adapt to the compact, information-dense nature of short-form videos. These methods generally lack support for multiple controllable styles and fail to align with platform-specific cultural and linguistic norms.

To address these limitations, we propose \textbf{LOLGORITHM}, a novel, extensible, modular multi-agent architecture for short-form video comment generation. The framework leverages MLLMs as its generation backbone and supports six controllable comment styles: puns, rhyming, memes, sarcasm, humor, and content extraction. The system comprises three core modules: \textit{video content summarization}, \textit{video content classification}, and \textit{comment generation}, with the video content summarization module also incorporating dataset construction functionality. In addition, we collect data from Douyin and YouTube to build a bilingual dataset covering five high-engagement video categories, enhancing contextual fit and community adaptability.

We further conduct ablation studies and model comparison experiments. Beyond comparing against baselines, we invoke multiple contemporary LLMs to assess the robustness of our framework. Evaluation combines automated metrics---namely originality, relevance, and style conformity---with human preference analysis. Results show that LOLGORITHM achieves leading performance on both platforms: across 40 videos evaluated by 105 respondents, user preference rates exceed 84.29\% on Douyin and reach 80.46\% on YouTube. The main contributions of this work are as follows:
\begin{enumerate}[leftmargin=*]
\item The first modular multi-agent architecture specifically designed for short-form video comment generation.
\item A comprehensive bilingual dataset for model training and evaluation.
\item Empirical validation of effective style control and multimodal context modeling for generating platform-aligned comments.
\end{enumerate}

%% file: input/related_work.tex
\section{Related Work}

\subsection{Video Summarization}
\citet{Apostolidis2021VideoSU} surveyed over 40 deep learning-based video summarization methods, encompassing supervised, weakly supervised, unsupervised, and multimodal approaches. They proposed a unified classification framework, analyzing strategies in terms of extraction efficiency, contextual modeling, and scene adaptability. \citet{Otani2022VideoSO} defined video summarization as extracting key information from long videos to improve viewing efficiency across domains, introducing a taxonomy based on Domains, Purposes, and Formats to guide standardization. \citet{Tonge2022ANA} developed a static summarization method using keyframe extraction and storyboard generation, improving narrative coherence and user engagement. \citet{Yao2022MultiLevelSN} proposed a multi-level spatiotemporal framework that jointly models frames, fragments, and shots, showing strong industrial applicability. Addressing data scarcity, \citet{Hua2024V2XumLLMCV} introduced Instruct-V2Xum, a dataset of 30{,}000 YouTube videos (40--940s) with frame-referenced summaries and a 16.39\% compression ratio, offering high semantic alignment for multimodal research.

\subsection{Live-Streaming Danmaku Generation}
Research on live-streaming danmaku generation has progressed from corpus construction to multimodal and personalized comment modeling. \citet{Ma2018LiveBotGL} pioneered the field with a Bilibili corpus and the LiveBot framework. \citet{Wang2020VideoICAV} introduced the denser VideoIC dataset and MML-CG, a multimodal multitask model surpassing LiveBot in accuracy and diversity. \citet{Sun2023ViCoEV} incorporated ``like'' counts in ViCo-20k to reflect human preferences. \citet{Fang2020Video2CommonsenseGC} broadened semantic scope via commonsense-grounded comment generation. \citet{Lalanne2023LiveChatVC} proposed a Triple Transformer Encoder using visual, audio, and textual contexts, alongside the English LiveChat dataset. \citet{Luo2024EngagingLV} introduced socially engaging comment generation with a multimodal dataset annotated with ``like'' signals. \citet{Wu2024UnderstandingHP} integrated user preference modeling and video segmentation to enhance alignment and anthropomorphism. \citet{Lin2024PersonalizedVC} proposed PerVidCom for personalized video comment generation, emphasizing style migration and semantic alignment. \citet{Gao2023LiveChatAL} focused on persona-driven interaction using a large-scale dialogue dataset with host speech, user comments, and profiles.

\subsection{Research Gap}
Despite notable progress in video summarization and live-streaming danmaku generation, directly applying these techniques to short-form video comment generation remains fundamentally challenging. First, summarization methods are designed to compress core content, and consequently tend to overlook user opinions, emotional cues, and social context---precisely the elements that make comments engaging and platform-relevant. Second, existing danmaku generation models rely heavily on historical behavior within single-user, single-stream contexts, which severely limits their generalization across users and content types, particularly in cross-platform scenarios. Such frameworks also exhibit poor robustness when confronted with the information-dense nature of short-form videos. Third, platform-specific architectural designs further hinder adaptability to the varying content styles, linguistic norms, and interaction patterns found across different short-video platforms. Finally, existing methods are predominantly monolingual, and directly prompting an LLM to generate comments introduces a distinct failure mode: the model produces stylistically formatted video summaries rather than authentic comments. Compounding these issues, no suitable dataset currently exists for short-form video comment generation, making both training and evaluation particularly difficult.

%% file: input/methodology.tex
\section{Methodology}

\begin{figure*}[t]
    \centering
    \includegraphics[width=\textwidth]{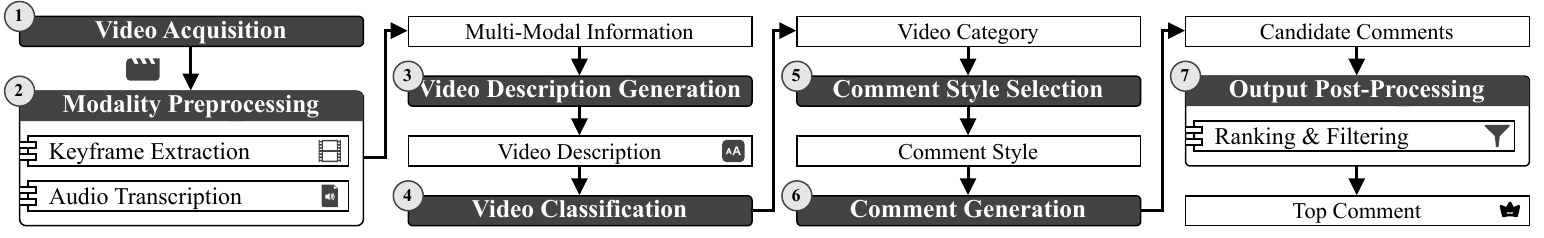}
    \caption{Workflow of the LOLGORITHM framework}
    \label{fig:framework}
    \Description{...}
\end{figure*}

LOLGORITHM is a novel, scalable, modular, multi-platform, and multilingual intelligent agent and framework for short-video comment generation. Currently, LOLGORITHM supports generating comments in both Chinese and English for TikTok and YouTube platforms, and is also capable of constructing dual-platform datasets. This section provides a detailed description of the framework through three components: the dataset generation module, the data preprocessing module, and the short-video comment generation module. Figure~\ref{fig:framework} presents the complete workflow of the LOLGORITHM framework.

\subsection{Dataset Construction Module}

We begin by introducing the dataset construction module. First, we manually selected 200 videos across five categories from two platforms and automatically crawled their comments via scripts, retaining the five most-liked comments per video. These comments were then manually annotated and classified, yielding a total of 1,000 labeled comments as the \textit{seed dataset}. This seed dataset was subsequently fed into the LOLGORITHM dataset generation module.

To ensure content diversity and comment interactivity, we performed balanced sampling across five video categories with high user engagement:

\begin{enumerate}[leftmargin=*]
    \item \textbf{Talk Shows}: Videos featuring hosts or speakers delivering conversational content to an audience.
    \item \textbf{Humorous Commentary}: Videos containing comedic narration or commentary on a wide range of topics.
    \item \textbf{Funny Animals}: Videos showcasing amusing animal behaviors and interactions.
    \item \textbf{Everyday Life Skits}: Videos depicting humorous situations drawn from daily life.
    \item \textbf{Comedy Short Dramas}: Scripted short-form comedic performances and sketches.
\end{enumerate}

The annotation process was carried out collaboratively by multiple researchers with expertise in linguistics and short-video content analysis, ensuring the accuracy and consistency of the labels. We categorize comment generation styles into six distinct types:

\begin{enumerate}[leftmargin=*]
    \item \textbf{Homophonic Wordplay (Puns)}: Exploiting homophones or near-homophones to create humor, a technique particularly prevalent in Chinese comments.
    \item \textbf{Rhyming}: Enhancing rhythmic flow through end rhymes, improving the readability and entertainment value of comments.
    \item \textbf{Meme Application}: Referencing internet memes or trending phrases to strengthen cultural resonance and shareability.
    \item \textbf{Irony (Sarcasm)}: Conveying viewpoints through sarcasm or irony, commonly found in teasing or critical comments.
    \item \textbf{General Humor}: Expressing humor directly without relying on linguistic devices or cultural references.
    \item \textbf{Content Extraction}: Commenting by directly referencing information from the video content, reflecting semantic alignment between the comment and the video.
\end{enumerate}

The LOLGORITHM dataset generation module adopts a six-stage end-to-end pipeline that automatically transforms raw short videos into a semantically annotated multimodal comment dataset. In \textbf{Stage 1}, the system automatically selects videos meeting the specified criteria on the corresponding platform based on user-defined content labels and topic tags. In \textbf{Stage 2}, the playback URL and description text of each video are extracted and the video file is downloaded. 

In \textbf{Stage 3}, all comments are collected and sorted in descending order by like count, with the top five retained as high-quality comment samples. In \textbf{Stage 4}, the system uniformly samples frames from each video and extracts the audio track, which is then fed into the Speech transcription model for automatic speech recognition to obtain a textual transcription of the video's spoken content. 

In \textbf{Stage 5}, the sampled frame sequences are batched together 
with the audio transcription and video tags, and a content description 
of each video is generated via a local MCP server that orchestrates the 
entire pipeline. The server exposes six tools to the agent: fetching the 
next pending video, assembling its visual and textual context, saving the generated description, and handling retries or failures. Notably, since Lolgorithm supports both locally deployed models and API-based token consumption, the agent first applies a tiered sampling strategy to select $k$ representative frames from a video containing $N$ total frames:

\begin{equation}
k = \begin{cases}
N  & \text{if } N \leq 12 \\
12 & \text{if } 12 < N \leq 60 \\
16 & \text{if } 60 < N \leq 160 \\
24 & \text{if } N > 160
\end{cases}
\end{equation}

\begin{figure*}[t]
    \centering
    \includegraphics[width=\textwidth]{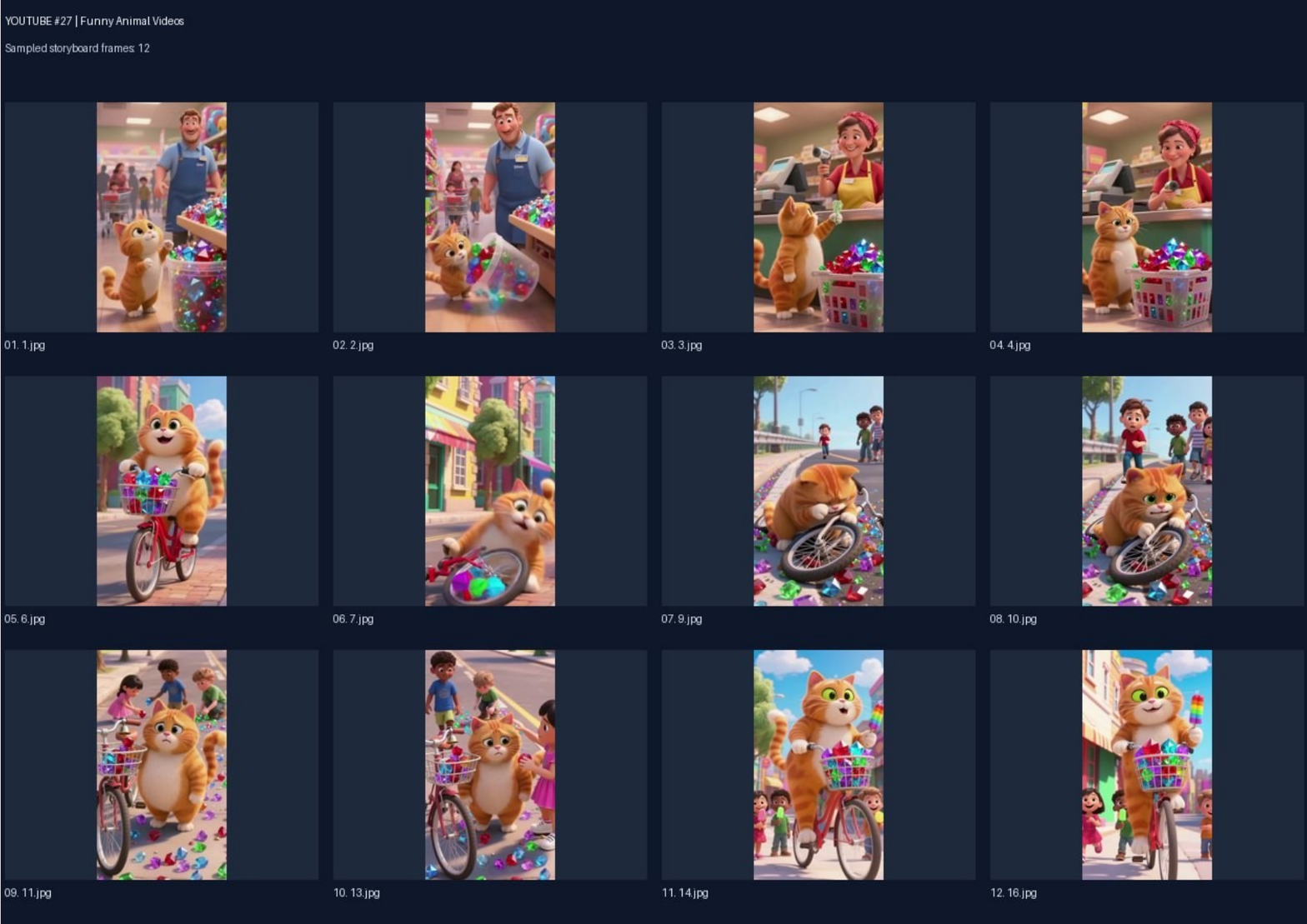}
    \caption{A Single Composite Image Example}
    \label{fig:frameCombine}
    \Description{...}
\end{figure*}

\noindent The $k$ frames are drawn by partitioning $N$ into $k$ 
equal-width buckets and selecting the midpoint index of each bucket, 
ensuring uniform temporal coverage. These frames are then stitched into 
a single composite image before being sent to the model, as illustrated 
in Figure~\ref{fig:frameCombine}. This design neither compromises the 
quality of the generated descriptions nor affects the semantic 
understanding of the visual content, while simultaneously reducing the 
number of API calls and conserving token usage, thereby substantially 
accelerating processing speed. 

\textbf{Stage 6} performs semantic label annotation (C\_label) on each of the top-five comments using a three-tier cascaded decision strategy. The first tier applies a prioritized regular-expression rule set for keyword matching; if a match is found, the corresponding label is assigned directly. If no rule is triggered, the second tier computes the TF-IDF cosine similarity between the comment and the video description. Specifically, both the comment $c$ and the video description $D_i$ are represented as high-dimensional sparse vectors weighted by Term Frequency–Inverse Document Frequency (TF-IDF), where each dimension corresponds to a token and its weight reflects how distinctive that token is relative to the entire corpus. The semantic proximity between the two texts is then measured by the cosine of the angle between their respective vectors:

\begin{equation}
    \text{sim}(c, D_i) = \frac{\vec{v}_c \cdot \vec{v}_{D_i}}{\|\vec{v}_c\| \cdot \|\vec{v}_{D_i}\|}
    \label{eq:cosine}
\end{equation}

\noindent where the numerator is the dot product of the two vectors, and the denominator is the product of their $\ell_2$ norms. The resulting value lies in $[0, 1]$; a value closer to 1 indicates that the comment is semantically close to the video description, implying that the comment is likely extracted from or directly referencing the video content. When $\text{sim}(c, D_i) \geq 0.10$, the comment is labeled as \textit{Content Extraction}. For comments that remain unlabeled after the second tier, the third tier first attempts to map them to a semantic category via an emotion lexicon. If the emotion-based mapping is also inconclusive, the system falls back to a TF-IDF-weighted $k$-Nearest Neighbor ($k$-NN) vote, finding the $k=5$ most similar comments in the training set and determining the label by majority vote. When the similarity of even the nearest neighbor falls below a minimum threshold, the system further degrades to a maximum a posteriori (MAP) estimate conditioned on the video label:

\begin{equation}
    \hat{y}_c = \arg\max_{\ell} \; P(\ell \mid \text{video\_label})
    \label{eq:map}
\end{equation}

\noindent where $\ell$ ranges over all possible C\_label categories, and $P(\ell \mid \text{video\_label})$ is estimated empirically as the relative frequency of label $\ell$ among all annotated comments belonging to the same video category. In other words, when no reliable local evidence is available, the system assigns the most frequently observed comment label within that video category as the prediction, ensuring that the annotation remains informed by category-level priors rather than defaulting to an arbitrary guess.

This module ensures both the quality and the scale of the resulting dataset, yielding a bilingual corpus comprising 3267 videos and 16335 comments. Moreover, owing to the modular and fully automated nature of the pipeline, the dataset is inherently extensible: it can readily be applied to collect and annotate videos beyond the original five categories, supporting future expansion without requiring manual redesign of the annotation workflow. 

\subsection{Video Processing Module}

Next, we introduce the video processing module. This module is responsible for processing the target videos for which comments are to be generated. Users can flexibly specify their input according to their own needs: they may provide a set of topic tags and allow LOLGORITHM to automatically search for matching videos on the corresponding platform, or they may directly supply the URLs of specific videos for LOLGORITHM to process. 
In \textbf{Stage 1}, the system automatically retrieves videos that satisfy the specified criteria based on the user-provided topic tags, or alternatively locates specific videos based on the URLs supplied by the user. 
In \textbf{Stage 2}, LOLGORITHM performs frame extraction on each video. To enable the model to better capture the key moments of a video during comment generation, we incorporate an audio-visual climax detection mechanism into this stage. The primary motivation for this design lies in the fundamental characteristics of short-form video content: unlike conventional long-form videos or live streams, short videos are typically under two minutes in duration. To attract and retain viewers within such a limited time frame, creators frequently employ sharp changes in lighting intensity and audio amplitude as attention-grabbing signals. We therefore adopt audio-visual climax detection to identify and isolate the high-intensity segments of each video. Frame sampling is performed at differentiated rates depending on the detected region: non-climax segments are sampled at one frame every two seconds, while climax segments are sampled at a substantially higher rate of five frames per second, ensuring that the most visually and acoustically salient moments are densely represented in the input to the downstream model. Formally, let $\mathcal{F}_{\text{normal}}$ and $\mathcal{F}_{\text{climax}}$ denote the frame sets sampled from non-climax and climax regions respectively, with sampling rates $r_{\text{normal}} = 0.5\,\text{fps}$ and $r_{\text{climax}} = 5\,\text{fps}$. The complete frame set for a video is then:

\begin{equation}
    \mathcal{F} = \mathcal{F}_{\text{normal}} \cup \mathcal{F}_{\text{climax}}
    \label{eq:frame_sampling}
\end{equation}

Concurrently, the audio track is extracted and fed into a speech recognition model to obtain a full textual transcription $T$ of the spoken content in the video.

In \textbf{Stage 3}, the sampled frames are stitched together in temporal order into a single composite image. This composite image, along with the audio transcription $T$ and the video topic tag, is jointly input into a Multimodal Large Language Model (MLLM), which produces a comprehensive natural language description $D$ of the video content. Stitching the frames into a single image before submission serves a dual purpose: it preserves the temporal structure of the visual narrative while simultaneously reducing the number of API calls and conserving token consumption, thereby accelerating overall processing speed without compromising the quality of the generated description.

\subsection{Short-Video Comment Generation Module}

The short-video comment generation module receives the output of the data preprocessing module and processes each target video through four sequential stages: semantic retrieval, comment style decision, hot meme augmentation, and prompt construction, ultimately driving a large language model to generate natural and authentic comments.

\textbf{Semantic Retrieval.} The video introduction, video description, and audio transcription of the target video $v$ are concatenated into a unified text representation, which is then encoded by an embedding model into a dense vector $\vec{e}_v$. If the category label $\ell_v$ of the target video is known, retrieval is performed preferentially within the same-category candidate space $\mathcal{I}(\ell_v)$, i.e., within the dataset; otherwise, the search degrades to a global search over all samples. Candidates are ranked by cosine similarity, and the top-$k$ most similar learning samples (default $k=3$) are selected:

\begin{equation}
    \mathcal{S}_v^* = \underset{s \in \mathcal{I}(\ell_v)}{\text{top-}k} \; \frac{\vec{e}_v \cdot \vec{e}_s}{\|\vec{e}_v\| \cdot \|\vec{e}_s\|}
    \label{eq:semantic_retrieval}
\end{equation}

\textbf{Comment Style Decision.} The C\_label distribution across all comments in the retrieved top-$k$ samples is aggregated, and the comment style type $\hat{c}$ to be generated for the current video is determined by majority voting:

\begin{equation}
    \hat{c} = \underset{c}{\arg\max} \sum_{s \in \mathcal{S}_v^*} \mathbf{1}[\text{C\_label}(s) = c]
    \label{eq:style_decision}
\end{equation}

\noindent Real comments corresponding to $\hat{c}$ are simultaneously extracted from the dataset as few-shot examples, with each sample contributing at most 2 examples, collectively forming the style reference set.

\textbf{Hot Meme Augmentation.} This stage is triggered only when $\hat{c}$ is of the ``Meme Application'' type. The system extracts keywords from the textual content of the target video and sequentially searches for the name and definition of relevant memes in the local meme cache and online meme encyclopedias (using \textit{Regeng Baike} for Chinese content, and \textit{Urban Dictionary} and \textit{Know Your Meme} for English content). Retrieved meme information is persisted to the local cache for future reuse. Upon successful application of a meme, the newly generated comment is appended to the meme's expression field, thereby forming a self-growing meme knowledge base that continuously enriches with each use.

\textbf{Prompt Construction and Comment Generation.} The video introduction, MLLM-generated video description, audio transcription, comment style type $\hat{c}$, style reference examples, and meme information (if any) are assembled into a complete prompt according to a structured template, which is then fed into a local large language model. To balance generation diversity and quality, the system configures the temperature parameter $T = 0.75$, nucleus sampling parameter $\text{top-}p = 0.9$, and repetition penalty coefficient $r = 1.1$, ultimately producing a comment that is stylistically natural, contextually aligned with the current video, and consistent with the predetermined semantic type.

%% file: input/evaluation.tex
\section{Evaluation}

We implemented LOLGORITHM in approximately 5,000 lines of Python code, with Qwen3.5-9B serving as the MLLM backbone of the agent. The evaluation is designed to answer two research questions:

\noindent\textbf{RQ1:} Can LOLGORITHM generate comments similar to high-quality human-written comments?

\noindent\textbf{RQ2:} How significantly does the underlying model affect the overall performance and behavior of LOLGORITHM?

\noindent\textbf{RQ3:} How does LOLGORITHM perform compared to existing approaches in terms of human preferences?

\subsection{Evaluation Setup}

\subsubsection{Baseline Techniques}

We designed two ablation experiments to verify the contribution of each core component in LOLGORITHM, and to explore the impact of the underlying model choice on it. 

\textbf{(1)} We first provide only the video content description directly to the model and generate comments using the same prompt as LOLGORITHM, in order to verify the contribution of the comment generation module to LOLGORITHM.

\textbf{(2)} To further account for the influence of the underlying model on agent performance, in addition to Qwen3.5-9b, we separately generate comments using three additional models --- GLM-5.4, DeepSeek-R1, and LLaMa4 --- enabling fair cross-model comparison under identical experimental conditions and assessing the impact of the backbone LLM choice on final comment quality.

We also compare LOLGORITHM against two representative baseline methods: \textbf{(1) V2Xum-LLM}~\citep{Hua2024V2XumLLMCV} — a multimodal summarization model representing the content summarization paradigm; and \textbf{(2) LiveChat}\citep{Gao2023LiveChatAL} — a real-time interactive comment generation system designed for live-streaming scenarios, representing the real-time interaction paradigm. Comparisons with these two categories of methods allow us to comprehensively evaluate the advantages of LOLGORITHM over existing approaches on the short-video comment generation task.

\subsubsection{Benchmark Videos}

We constructed a dataset of 40 short videos crawled from Douyin (Chinese) and YouTube (English), with 20 videos from each platform.
These short videos are different from those used in the tool's knowledge base.
It is worth noting that we purposefully included several videos that do not fall into the categories of the tool's video classification to increase the challenge and simulate real-world scenarios, thereby evaluating how the tool performs on previously unseen types of videos.
Table~\ref{tab:videos} shows the detailed information of the videos used for benchmarking.
In addition to these videos, we also used the dataset construction module to collect five highly-rated comments for each video as reference standards for answering RQ1, and created a benchmark dataset for scoring in the automatic scoring framework.

\begin{table}[t]
\centering
\caption{Details of the videos used for benchmarking. TS: Talk Show; HC: Humorous Commentary; FA: Funny Animal; DLJ: Daily Life Jokes; CS: Comedy Skits.}
\label{tab:videos}
\begin{tabular}{lcccccc}
\hline
\textbf{Platform} & \textbf{TS} & \textbf{HC} & \textbf{FA} & \textbf{DLJ} & \textbf{CS} & \textbf{Other} \\ 
\hline
YouTube  & 3 & 3 & 3 & 4 & 3 & 4 \\ 
Douyin   & 3 & 4 & 2 & 4 & 3 & 4 \\ 
\hline
\end{tabular}
\end{table}

\subsubsection{Evaluation Metrics}

To comprehensively assess the performance of LOLGORITHM, we adopt a dual evaluation strategy combining automatic scoring and human preference analysis. 
This approach ensures our system is benchmarked both computationally and socially, capturing both objective metrics and subjective user engagement potential.

\paragraph{Automatic Scoring Framework.}
We design an automatic scoring system that evaluates generated comments along three dimensions: originality ($S_o$), relevance ($S_r$), and style conformity ($S_s$). Each dimension has a maximum score of 10 points, and the final score is:
\begin{equation}
S_{\text{total}} = \frac{S_o + S_r + S_s}{3}
\end{equation}
where $S_{\text{total}}, S_o, S_r, S_s \in [0,10]$.

\textbf{Originality ($S_o$):} 
We evaluate originality by comparing the similarity between each generated comment $c_i$ and existing comments. Let $\mathcal{D}_b$ and $\mathcal{D}_t$ denote benchmark and training datasets respectively, and $\mathcal{D} = \mathcal{D}_b \cup \mathcal{D}_t$. We compute:
\begin{equation}
m = \max\left(\max_{d \in \mathcal{D}} \text{sim}(c_i, d), \text{sim}(c_i, v_i)\right)
\end{equation}
\begin{equation}
S_o = 10 \cdot (1 - m)
\end{equation}
where $m$ is the maximum similarity score and $v_i$ is the target video content.

\textbf{Relevance ($S_r$):} 
Relevance measures appropriate connection with video content. We first compute a baseline $b$ for human comments:
\begin{equation}
b = \frac{1}{|\mathcal{D}_b|} \sum_{(c_j, v_j) \in \mathcal{D}_b} \text{sim}(c_j, v_j)
\end{equation}
For a generated comment $c_i$ and video $v_i$:
\begin{equation}
S_r = 10 \cdot \exp\left(-\frac{|\text{sim}(c_i, v_i) - b|^2}{2\sigma^2}\right)
\end{equation}
where $\sigma$ is a scaling parameter.

\textbf{Style Conformity ($S_s$):} 
Style conformity combines length and sentiment matching. We found human-written English comments typically contain 63--72 words, while Chinese comments contain 25--35 characters. Using DistilBERT~\citep{Sanh2019DistilBERTAD} for English and BERT-Chinese~\citep{Devlin2019BERTPO} for sentiment analysis:
\begin{equation}
S_s = S_l + S_{st}
\end{equation}
where:
\begin{equation}
S_l = \begin{cases} 
5 & L_{\min} \leq |c_i| \leq L_{\max} \\
5e^{-\frac{(|c_i| - L_n)^2}{2\sigma_L^2}} & \text{else}
\end{cases}
\end{equation}
with $L_{\min}$, $L_{\max}$ as bounds and $L_n$ as nearest bound, and:
\begin{equation}
S_{st} = \begin{cases} 
5 & \text{sent}(c_i) = \text{sent}(v_i) \\
0 & \text{else}
\end{cases}
\end{equation}

\paragraph{Human Preference Analysis.}
We conducted a manual evaluation using a multimodal anonymous questionnaire and this experiment has obtained ethical review approval. Each question consisted of a target video and four generated comments: one from each of the three baseline models and one from LOLGORITHM. Respondents watched the video and selected the comment they were most likely to engage with, simulating real-world platform interaction. Two questionnaires were administered to native speakers aged 20--50.

\subsection{RQ1: Can LOLGORITHM generate comments similar to high-quality human-written comments?}

To answer this research question, we conducted quantitative analysis of LOLGORITHM and baseline methods through our automatic evaluation framework. Table~\ref{tab:douyin_results} and Table~\ref{tab:youtube_results} present detailed scoring results on the Douyin and YouTube datasets, respectively.

\begin{table}[t]
\centering
\caption{Scoring results on Douyin dataset.}
\label{tab:douyin_results}
\begin{tabular}{lcccc}
\hline
\textbf{Model} & \textbf{Orig.} & \textbf{Rel.} & \textbf{Style} & \textbf{Total} \\
\hline
V2Xum-LLM & 1.38 & \textbf{8.24} & 1.10 & 3.57 \\
LiveChat & 1.21 & 7.49 & 5.44 & 4.71 \\
Qwen3.5-9b (Direct) & 1.75 & 7.66 & 2.42 & 3.94 \\
\textbf{LOLGORITHM} & \textbf{1.75} & 7.98 & \textbf{6.59} & \textbf{5.44} \\
\hline
\end{tabular}
\end{table}

\begin{table}[t]
\centering
\caption{Scoring results on YouTube dataset.}
\label{tab:youtube_results}
\begin{tabular}{lcccc}
\hline
\textbf{Model} & \textbf{Orig.} & \textbf{Rel.} & \textbf{Style} & \textbf{Total} \\
\hline
V2Xum-LLM & 2.73 & \textbf{8.03} & 1.75 & 4.17 \\
LiveChat & 2.37 & 5.14 & \textbf{6.77} & 4.76 \\
Qwen3.5-9b (Direct) & 2.21 & 5.40 & 4.09 & 3.90 \\
\textbf{LOLGORITHM} & \textbf{3.26} & 7.70 & 4.76 & \textbf{4.90} \\
\hline
\end{tabular}
\end{table}

As shown in Table~\ref{tab:douyin_results}, on the Douyin dataset, LOLGORITHM achieved the highest total score of 5.44, outperforming all baseline methods. In the relevance dimension, LOLGORITHM scored 7.98, surpassing LiveChat (7.49) and Qwen3.5-9b Direct (7.66), and approaching V2Xum-LLM's leading score of 8.24, demonstrating its strong ability to maintain topical alignment with video content. More notably, LOLGORITHM achieved the highest style conformity score of 6.59, substantially outperforming V2Xum-LLM (1.10) and Qwen3.5-9b Direct (2.42), and exceeding LiveChat (5.44), indicating that its generated comments closely match the linguistic patterns, length requirements (25--35 characters for Chinese), and sentiment expectations of the Douyin platform. In terms of originality, LOLGORITHM tied with Qwen3.5-9b Direct at 1.75, both leading over V2Xum-LLM (1.38) and LiveChat (1.21). The balanced performance across all three dimensions highlights LOLGORITHM's ability to simultaneously achieve strong relevance, high style conformity, and competitive originality --- a balance that neither V2Xum-LLM nor LiveChat was able to attain.

On the YouTube dataset (Table~\ref{tab:youtube_results}), LOLGORITHM achieved the best overall performance with a total score of 4.90, outperforming LiveChat (4.76), V2Xum-LLM (4.17), and Qwen3.5-9b Direct (3.90). LOLGORITHM demonstrated the highest originality score of 3.26, surpassing V2Xum-LLM (2.73), LiveChat (2.37), and Qwen3.5-9b Direct (2.21), showcasing its ability to generate distinctive and creative comments. In the relevance dimension, LOLGORITHM scored 7.70, significantly exceeding LiveChat (5.14) and Qwen3.5-9b Direct (5.40), and only marginally behind V2Xum-LLM's 8.03, reflecting its strong contextual understanding of video content. While LiveChat achieved the highest style conformity score (6.77), its poor relevance score (5.14) resulted in a lower total (4.76). In contrast, LOLGORITHM's style conformity of 4.76, combined with its strong relevance and leading originality, enabled it to achieve the highest overall score. This pattern also holds for V2Xum-LLM, which despite its high relevance (8.03), scored only 1.75 in style conformity, yielding the lowest total score of 4.17 among all methods.

Overall, the automatic evaluation results demonstrate that LOLGORITHM consistently achieves the highest total scores across both platforms, validating the effectiveness of its multi-agent collaborative framework. Its ability to balance all three evaluation dimensions --- maintaining strong relevance (7.98 on Douyin, 7.70 on YouTube), achieving competitive or leading originality (1.75 on Douyin, 3.26 on YouTube), and delivering superior style conformity (6.59 on Douyin, 4.76 on YouTube) --- confirms that LOLGORITHM can generate comments that closely resemble high-quality human-written comments while adapting to the distinct cultural and stylistic norms of each platform.

\subsection{RQ2: How significantly does the underlying model affect the overall performance and behavior of LOLGORITHM?}

To investigate how the choice of backbone large language model affects comment generation quality, we evaluated four models --- Qwen3.5-9b (LOLGORITHM), DeepSeek-R1, LLaMa4, and GPT-5.4 --- under identical experimental conditions on both platforms. To further isolate the contribution of the framework itself, Qwen3.5-9b (Direct) will also serves as a ablation baseline, where the same model generates comments without the multi-agent framework. Table~\ref{tab:model_comparison_douyin} and Table~\ref{tab:model_comparison_youtube} present the results.

\begin{table}[t]
\centering
\caption{Impact of backbone LLM on Douyin dataset.}
\label{tab:model_comparison_douyin}
\begin{tabular}{lcccc}
\hline
\textbf{Model} & \textbf{Orig.} & \textbf{Rel.} & \textbf{Style} & \textbf{Total} \\
\hline
Qwen3.5-9b (Direct) & 1.75 & 7.66 & 2.42 & 3.94 \\
\hline
DeepSeek-R1 & 1.22 & \textbf{8.33} & 4.37 & 4.64 \\
LLaMa4      & 1.29 & 8.26 & 5.43 & 4.99 \\
LOLGORITHM (Qwen3.5-9b) & \textbf{1.75} & 7.98 & 6.59 & 5.44 \\
\textbf{GPT-5.4} & 1.43 & 8.25 & \textbf{7.45} & \textbf{5.71} \\
\hline
\end{tabular}
\end{table}

\begin{table}[t]
\centering
\caption{Impact of backbone LLM on YouTube dataset.}
\label{tab:model_comparison_youtube}
\begin{tabular}{lcccc}
\hline
\textbf{Model} & \textbf{Orig.} & \textbf{Rel.} & \textbf{Style} & \textbf{Total} \\
\hline
Qwen3.5-9b (Direct) & 2.21 & 5.40 & 4.09 & 3.90 \\
\hline
LOLGORITHM (Qwen3.5-9b) & 2.25 & 7.70 & 4.76 & 4.90 \\
GPT-5.4     & 2.27 & 7.78 & 4.73 & 4.93 \\
LLaMa4      & 2.09 & 7.73 & 5.49 & 5.11 \\
\textbf{DeepSeek-R1} & \textbf{2.32} & \textbf{8.04} & 5.10 & \textbf{5.15} \\
\hline
\end{tabular}
\end{table}

The most striking finding is not the variation among different backbone models, but rather the substantial and consistent improvement that the LOLGORITHM framework delivers over direct generation using the same underlying model. On the Douyin dataset, Qwen3.5-9b (Direct) scored only 3.94, while LOLGORITHM built on the identical Qwen3.5-9b backbone achieved 5.44 --- an improvement of 1.50 points, representing a 38.1\% relative gain. This improvement is most pronounced in style conformity, which rose from 2.42 to 6.59, a gain of 4.17 points, demonstrating that the framework's platform-specific prompt design and example-guided generation are critical for producing comments that conform to Douyin's concise and culturally distinctive style. On the YouTube dataset, a similarly significant improvement is observed: Qwen3.5-9b (Direct) scored 3.90, while LOLGORITHM achieved 4.90, a gain of 1.00 point (25.6\% relative improvement), with relevance increasing markedly from 5.40 to 7.70 --- a gain of 2.30 points --- attributable to the framework's structured retrieval and category-aware example selection mechanisms.

In contrast, the performance variation introduced purely by swapping the backbone model --- while holding the framework constant --- is substantially smaller. On Douyin, the total scores of the four framework-equipped models range from 4.64 to 5.71, a spread of 1.07 points. On YouTube, this spread narrows further to just 0.25 points (4.90 to 5.15), with all four models converging to a similar level of quality. This tight clustering across models with fundamentally different architectures and training paradigms strongly suggests that the LOLGORITHM framework, rather than the backbone model's intrinsic capability, is the primary driver of comment generation quality.

These findings lead to a clear conclusion: the multi-agent collaborative architecture of LOLGORITHM --- encompassing structured retrieval, category-aware example selection, and platform-specific prompt design --- provides a powerful and model-agnostic scaffold that elevates comment generation quality far beyond what any backbone model can achieve alone. The backbone LLM plays a secondary and non-decisive role; it is the framework's design that fundamentally determines the system's effectiveness, and this robustness ensures reliable high-quality performance regardless of which underlying model is deployed.

\begin{figure}[t]
  \centering
  \begin{minipage}{\linewidth}
    \centering
    \includegraphics[width=0.8\linewidth]{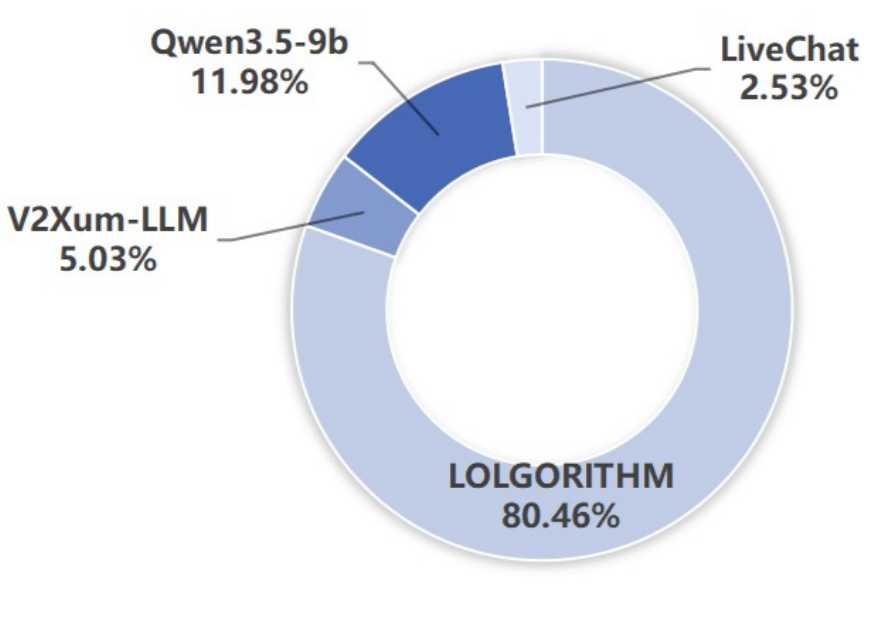}
    \Description{Pie chart showing YouTube human evaluation results.}
    \\(a) YouTube platform results.
  \end{minipage}
  \hfill
  \begin{minipage}{\linewidth}
    \centering
    \includegraphics[width=0.8\linewidth]{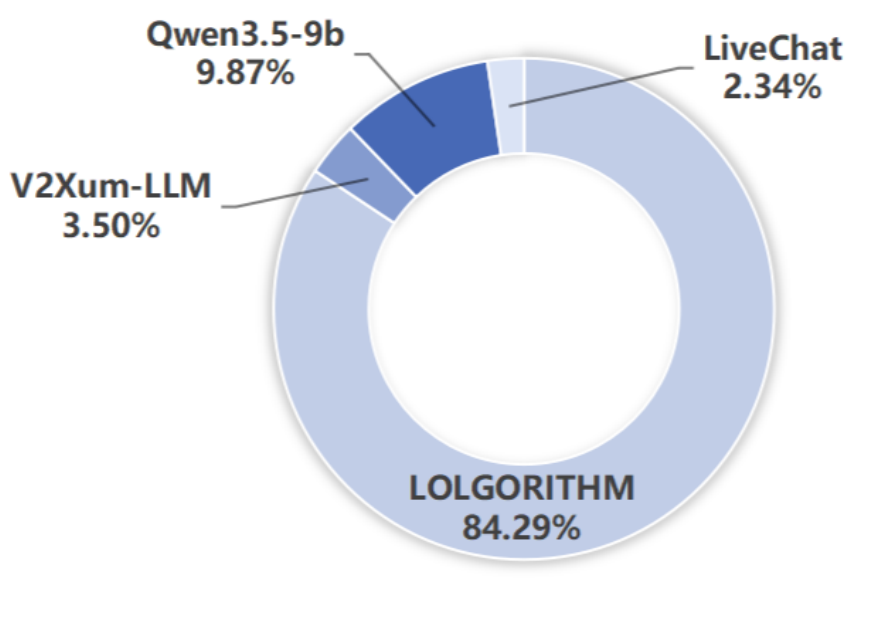}
    \Description{Pie chart showing Douyin human evaluation results.}
    \\(b) Douyin platform results.
  \end{minipage}
  \caption{Human evaluation results for comment generation.}
  \label{fig:human_eval_combined}
\end{figure}

\subsection{RQ3: How does LOLGORITHM perform compared to existing approaches in terms of human preferences?}

To evaluate LOLGORITHM's performance in terms of real user preferences, we conducted a large-scale human evaluation experiment. Figure~\ref{fig:human_eval_combined} presents the human preference evaluation results on both Douyin and YouTube platforms.

The human evaluation results provide compelling evidence of our system's effectiveness. On the YouTube platform, with 51 valid responses collected across 20 video samples, LOLGORITHM was selected as the most preferred comment in 80.46\% of cases, significantly outperforming other baseline methods: LiveChat received only 2.53\% selection rate, V2Xum-LLM achieved 5.03\%, and Qwen3.5-9b (Direct) obtained 11.98\%. On the Douyin platform, the results were even more remarkable: through 56 valid responses across 20 video samples, LOLGORITHM achieved an overwhelming 84.29\% selection rate, while all other methods performed below 10\% (LiveChat: 2.34\%, Qwen3.5-9b (Direct): 9.87\%, V2Xum-LLM: 3.5\%).

These results clearly demonstrate that LOLGORITHM has significant advantages in generating user-preferred comments, with its multi-agent collaborative framework better capturing the comment culture and user expectations of different platforms.

\subsection{Case Study}

\begin{figure}[t]
  \centering
  \begin{minipage}{\linewidth}
    \centering
    \includegraphics[width=\linewidth]{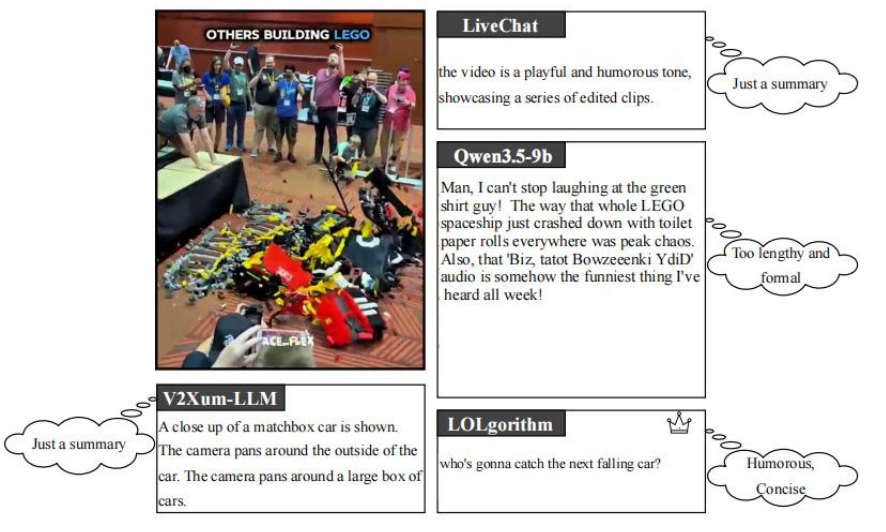}
    \Description{Example comments generated by different techniques on YouTube.}
    \\(a) YouTube example.
  \end{minipage}

  \begin{minipage}{\linewidth}
    \centering
    \includegraphics[width=\linewidth]{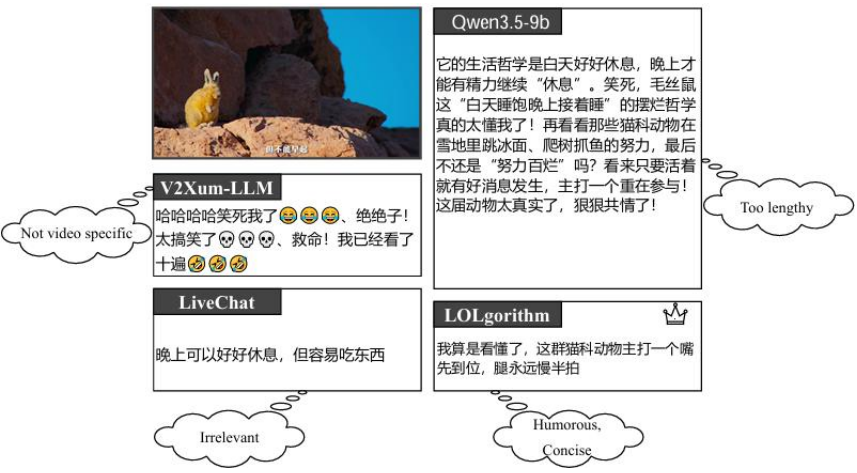}
    \Description{Example comments generated by different techniques on Douyin.}
    \\(b) Douyin example.
  \end{minipage}
  \caption{Example comments generated by different techniques.}
  \label{fig:examples}
\end{figure}

Figure~\ref{fig:examples} further illustrates LOLGORITHM's advantages over baseline methods through specific YouTube and Douyin video cases.

In the YouTube case, LOLGORITHM generated concise and humorous content that perfectly matches YouTube users' interactive style. In contrast, Qwen3.5-9b (Direct) was overly lengthy and formal, V2Xum-LLM only provided simple video descriptions, while LiveChat lacked specificity.

In the Douyin case, LOLGORITHM cleverly used wordplay and internet slang, embodying the humorous style favored by Douyin users. Qwen3.5-9b (Direct), while rich in content, was too lengthy, not fitting Douyin's culture of concise and entertaining comments. V2Xum-LLM, though containing emojis, lacked specific relevance to the video content, while LiveChat was completely irrelevant.

These cases clearly demonstrate LOLGORITHM's superior ability to understand platform-specific comment cultures and generate comments of appropriate length and style, explaining why it achieved such high selection rates in human preference evaluation.

%% file: input/conclusion.tex
\section{Conclusion}
In this paper, we presented LOLGORITHM, a modular multi-agent framework for stylized short-form video comment generation, addressing key limitations of existing video summarization and danmaku generation approaches that fail to capture platform-specific comment cultures and linguistic norms.


Experiments on bilingual YouTube and Douyin datasets demonstrate that LOLGORITHM consistently outperforms all baseline methods across both automatic scoring and human preference evaluation, with selection rates of 84.29\% on Douyin and 80.46\% on YouTube across 107 respondents. In automatic scoring, LOLGORITHM achieves the highest total scores on both platforms, with particular strengths in style conformity (6.59 on Douyin, 4.76 on YouTube) and relevance (7.98 on Douyin, 7.70 on YouTube). Ablation analysis further confirms that performance gains stem primarily from the multi-agent framework architecture rather than the backbone LLM, with backbone substitution causing only marginal variation (4.90\% to 5.15\% on YouTube), underscoring the robustness and model-agnostic nature of our approach.

Future work will explore expanding platform support and extending the comment style taxonomy to cover emerging internet trends.